\documentclass{article}

    \PassOptionsToPackage{square,numbers}{natbib}

\usepackage[preprint]{neurips_2021}

\usepackage[utf8]{inputenc} 
\usepackage[T1]{fontenc}    
\usepackage{hyperref}       
\usepackage{url}            
\usepackage{booktabs}       
\usepackage{amsfonts}       
\usepackage{nicefrac}       
\usepackage{microtype}      
\usepackage{xcolor}         

\usepackage{graphicx}
\usepackage{subfig}
\usepackage{algorithm}
\usepackage{gensymb}
\usepackage{algpseudocode}

\newcommand\blfootnote[1]{%
  \begingroup
  \renewcommand\thefootnote{}\footnote{#1}%
  \addtocounter{footnote}{-1}%
  \endgroup
}

\title{AutoProtoNet: Interpretability for Prototypical Networks}

\author{%
  Pedro~Sandoval-Segura \\
  Department of Computer Science\\
  University of Maryland\\
  \texttt{psando@cs.umd.edu} \\
   \And
   Wallace Lawson  \\
   Navy Center for Applied Research in Artificial Intelligence \\
   Naval Research Laboratory \\
   \texttt{ed.lawson@nrl.navy.mil} \\
}

\begin{document}

\maketitle

\begin{abstract}
In meta-learning approaches, it is difficult for a practitioner to make sense of what kind of representations the model employs. Without this ability, it can be difficult to both understand what the model knows as well as to make meaningful corrections. To address these challenges, we introduce AutoProtoNet, which builds interpretability into Prototypical Networks by training an embedding space suitable for reconstructing inputs, while remaining convenient for few-shot learning. We demonstrate how points in this embedding space can be visualized and used to understand class representations. We also devise a prototype refinement method, which allows a human to debug inadequate classification parameters. We use this debugging technique on a custom classification task and find that it leads to accuracy improvements on a validation set consisting of in-the-wild images. We advocate for interpretability in meta-learning approaches and show that there are interactive ways for a human to enhance meta-learning algorithms.

\end{abstract}

\section{Introduction}
\blfootnote{DISTRIBUTION STATEMENT A. Approved for public release: distribution unlimited.}

It is expensive and time-consuming to collect data to train current state-of-the-art image classification systems \cite{kolesnikov2019bit}. When a classification algorithm is deployed, new classes or labels cannot be easily added without incurring new costs related to re-training the model \cite{altae2016data}\cite{bengio2012deep}. Meta-learning approaches for few-shot learning solve both these problems by training networks that learn quickly from little data with computationally inexpensive fine-tuning \cite{vinyals2016matching}\cite{snell2017prototypical}\cite{lee2019metaoptnet}. Despite these methods performing well on benchmark few-shot image classification tasks, these methods are not interpretable; a human may have no way of knowing why a certain classification decision was made. Additionally, the lack of interpretability limits any kind of debugging of network representations. In this work, we take a step toward the development of a meta-learning algorithm which can learn in a few-shot setting, can handle new classes at test time, is interpretable enough for a human to understand how the model makes decisions, and which can be debugged in a simple way.

We revisit Prototypical Networks (ProtoNets) \cite{snell2017prototypical} as the focus of our study. ProtoNets are based on a simple idea: there exists an embedding space where images cluster around a single ``prototype'' for each class. Given the simplicity of this few-shot learning approach, it makes sense to ask: what does a prototype look like? And, have we learned an adequate prototype representation?

The outcomes of our study can be summarized as follows:
\begin{itemize}
    \item We introduce AutoProtoNet, which merges ideas from autoencoders and Prototypical Networks, to perform few-shot image classification and prototype reconstruction.
    \item We use AutoProtoNet to visualize prototypes and find that they are comparable in quality to those of an autoencoder. AutoProtoNet also remains accurate on few-shot image classification benchmarks.
    \item We devise a prototype refinement method, which can be used to debug inadequate prototypes, and we validate the performance of the resulting model using a novel validation set of in-the-wild images.
\end{itemize}

Our goal in this work is to elucidate the benefits of learning embeddings that can be visualized and interpreted by humans. To the best of our knowledge, there is no meta-learning approach that allows for a human to play a role in the fine-tuning of the base model. 

\section{Related Work}

\subsection{Meta-learning and Prototypical Networks}
\label{subsection:meta-learning}

Before meta-learning, transfer learning was used to handle few-shot problems. In transfer learning, a feature extractor is trained on a large dataset, then fine-tuned for new tasks \cite{bengio2012deep}. However, transfer learning has some drawbacks. For example, adding a new class may require re-training the model and, in the few-shot setting, overfitting few example images is possible. 

Meta-learning algorithms aim to learn a ``base'' model that can be quickly fine-tuned for a new task. The base model is trained using a set of training tasks $\{\mathcal{T}_i\}$, sampled from some task distribution. Each task consists of \emph{support} data, $\mathcal{T}_i^s$, and \emph{query} data, $\mathcal{T}_i^q$. Support data is used to fine-tune the model, while query data is used to evaluate the resulting model. Practically speaking, each task is an image classification problem involving only a small number of classes. The number of examples per class in the support set is called the \emph{shot}, and the number of classes is called the \emph{way}. For example, in 5-way 1-shot learning, we are given 1 example for each of the 5 classes to use for fine-tuning.


Following the meta-learning framework presented in \cite{goldblum2019robust}, Algorithm~\ref{alg:meta-learning-framework} can be used as a general way to understand both metric-learning methods \cite{vinyals2016matching} \cite{snell2017prototypical} and gradient-based methods like MAML \cite{finn2017maml}. 

\renewcommand{\algorithmicrequire}{\textbf{Input: }}
\renewcommand{\algorithmicensure}{\textbf{Output: }}
\begin{algorithm}
\caption{The meta-learning framework}
\label{alg:meta-learning-framework}
    \algorithmicrequire Base model, $F_{\theta}$ \\
    \algorithmicrequire Fine-tuning algorithm, $A$ \\
    \algorithmicrequire Learning rate, $\gamma$ \\
    \algorithmicrequire Distribution over tasks, $p(\mathcal{T})$
    \begin{algorithmic}[1]
        \State Initialize $\theta$, the weights of $F$ 
        \While{not done}
            \State Sample batch of tasks $\{\mathcal{T}_i\}_{i=1}^{n}$, where $\mathcal{T}_i \sim p(\mathcal{T})$ and $\mathcal{T}_i = (\mathcal{T}_i^s, \mathcal{T}_i^q)$
            \For{i=1,...,n}
                \State $\theta_i \gets A(\theta, \mathcal{T}_i^s)$ \Comment{Fine-tune model on $\mathcal{T}_i^s$ (inner loop)} 
                \State $g_i \gets \nabla_{\theta}\mathcal{L}(F_{\theta_i}, \mathcal{T}_i^q)$
            \EndFor
            \State $\theta \gets \theta - \frac{\gamma}{n}\sum_{i}g_i$ \Comment{Update base model parameters (outer loop)}
        \EndWhile
    \end{algorithmic}
\end{algorithm}

For ProtoNets \cite{snell2017prototypical}, the base model $F_{\theta}: \mathbb{R}^D \rightarrow \mathbb{R}^M$ is an embedding network which takes an image $x \in \mathbb{R}^D$ as input and outputs an embedding vector of dimension $M$. Suppose, for example, we have a $K$-way task $\mathcal{T}_i = (\mathcal{T}_i^s, \mathcal{T}_i^q)$ where $\mathcal{T}_i^s = \{(x_{i,1}, y_{i,1}), (x_{i,2}, y_{i,2}), ..., (x_{i,N}, y_{i,N})\}$, and where $y_{i,j} \in \{1,...,K\}$. Additionally, let $S_k \subset \mathcal{T}_i^s$ denote the set of support examples of class $k$. Then, a prototypical network computes a prototype $p_k$ for each class $k$ by computing a class-wise mean of embedded support examples:

\begin{equation}
    p_k = \frac{1}{|S_k|} \sum_{(x,y) \in S_k} F_{\theta}(x)
\end{equation}

Thus, in the case of ProtoNets, the fine-tuning algorithm $A$ does not update model parameters $\theta$, but instead it computes a set of prototypes which the base model will use to classify query data. We can think of $A$ as a function taking both embedding network parameters $\theta$ and support data $\mathcal{T}_i^s$ and returning a tuple $\theta_i$ consisting of a set of prototypes and an unchanged set of model parameters; i.e., $A(\theta, \mathcal{T}_i^s) = ( \{p_k\}_{i=0}^{k}, \theta ) = \theta_i$. In this way, $F_{\theta_i}$ in Algorithm~\ref{alg:meta-learning-framework} refers to using the base model parameters $\theta$ and the set of prototypes $\{p_k\}_{i=0}^{k}$ during inference. Given a distance function $d: \mathbb{R}^M \times \mathbb{R}^M \rightarrow [0,\infty)$ and a query point $x$, a ProtoNet produces a distribution over classes based on a softmax over distances to the prototypes in embedding space:

\begin{equation}
    p_{\theta}(y=k | x) = \frac{\exp(-d(F_{\theta}(x), p_k))}{\sum_{k'} \exp(-d(F_{\theta}(x), p_{k'}))}
    \label{equation:protonet-likelihood}
\end{equation}

Training proceeds by minimizing the negative log-likelihood $\mathcal{L}(\theta) = -\log p_{\theta}(y=k | x)$ of the true class $k$ using SGD. Unfortunately, ProtoNet does not provide a way to understand the embedding space or visualize $p_k$ -- a problem we directly address in this work.

\subsection{Understanding Meta-learning Approaches}

Investigating the ability of meta-learning methods to adapt to new tasks has been the subject of numerous studies. The success of meta-learning approaches certainly seems to suggest that the representations learned by meta-learning must be different than those learned through standard training \cite{goldblum2020unraveling}. \citet{goldblum2020unraveling} find that meta-learned feature extractors outperform classically trained models of the same architecture and suggest that meta-learned features are qualitatively different from conventional features. While work has been done to understand how the meta-learning networks train \cite{huang2019genthroughvis}\cite{frosst2019analyzing}, there has been little to no focus on developing tools to interpret the meta-learned models.

\subsection{Interpretability in Convolutional Models}

In safety or security-critical applications, understanding why a classification system made a certain prediction is important. Just because a classification system is highly accurate, does not mean the network has learned the right kinds of features \cite{ilyas2019adversarial}. We believe that a system that can demonstrate its logic semantically or visually is more likely to be trusted and used. Being that a ProtoNet is primarily a convolutional neural network, it is appropriate to understand progress on interpretability of convolutional neural networks (CNN). 

There are many research branches within the umbrella of CNN interpretability including visualizations of intermediate network layers \cite{zeiler2013visunderstandcnn}\cite{Mahendran2014understanddeepimagerep}\cite{simonyan2014deep}\cite{springenberg2015striving}, diagnosis of CNN representations \cite{zhang2017growing}\cite{zhang2017interpretingcnnknowledge}, and building explainable models \cite{zhang2018interpretable}. In contrast to works which focus their attention on CNN layers and activations, we take a more specific approach in visualizing embedding space for ProtoNets.

\citet{zhang2018interpretable} propose a compelling method of modifying convolutional layers so that each filter learns to represent a particular object part, thus allowing for each filter to correspond to a semantically meaningful image feature. We believe there could be interesting work incorporating this technique into meta-learning approaches, but is not appropriate for a shallow embedding network like the one we employ for ProtoNets.

\subsection{Generative Models}

Work on Variational Prototyping Encoder (VPE) \cite{kim2019variational} is most similar to ours in that a meta-task is used to learn an embedding space suitable for both few-shot learning and unseen data representation. In contrast, we do not focus on the image translation task from real images to prototypes and instead focus our attention on visualizing prototypes for interpretability and refinement. 

There are also a number of works which investigate connections between autoencoder architectures and meta-learning, but which are not directly applicable for interpretability of few-shot image classification. For example, \citet{wu2018metalearning} propose the Meta-Learning Autoencoder (MeLA) framework which learns a recognition and generative model to transform a single-task model into one that can quickly adapt to new tasks using few examples. However, their framework is meant for the more general understanding of \emph{tasks} like physical state estimation and video prediction, as opposed to the image classification tasks which we focus on. Similarly, \citet{epstein2019jointautoencoders} develop a meta-learning framework consisting of joint autoencoders for the purpose of learning multiple tasks simultaneously, but this approach is tailored more for the field of multi-task learning.

\section{Algorithm}
\label{algorithm}

Our interpretability algorithm takes advantage of the simplicity of the ProtoNet classification method. In particular, a ProtoNet classifies query data according to the class of the prototype which the query data's embedding is nearest to, typically in Euclidean space. This classification method raises an obvious question: what does a prototype look like? To answer this question, we extend ProtoNets with a decoder to reconstruct images from embeddings.

\subsection{Data}

The CIFAR-FS dataset \cite{bertinetto2018meta} is a recent few-shot image classification benchmark consisting of all $100$ classes from CIFAR-100 \cite{Krizhevsky2009LearningML}. Classes are randomly split into 64, 16, and 20 for meta-training, meta-validation, and meta-testing respectively. Every class contains $600$ images of size $32 \times 32$.

The \emph{mini}ImageNet dataset \cite{vinyals2016matching} is another standard benchmark for few-shot image classification. It consists of 100 randomly chosen classes from ILSVRC 2012 \cite{deng2009imagenet}, which are split into 64, 16, and 20 classes for meta-training, meta-validation, and meta-testing respectively. For every class, there are $600$ images of size $84 \times 84$. We adopt the commonly-used Ravi and Larochelle split proposed in \cite{ravi2017OptimizationAA}.

\subsection{Architecture}

\begin{figure}
  \centering
  \includegraphics[scale=0.20]{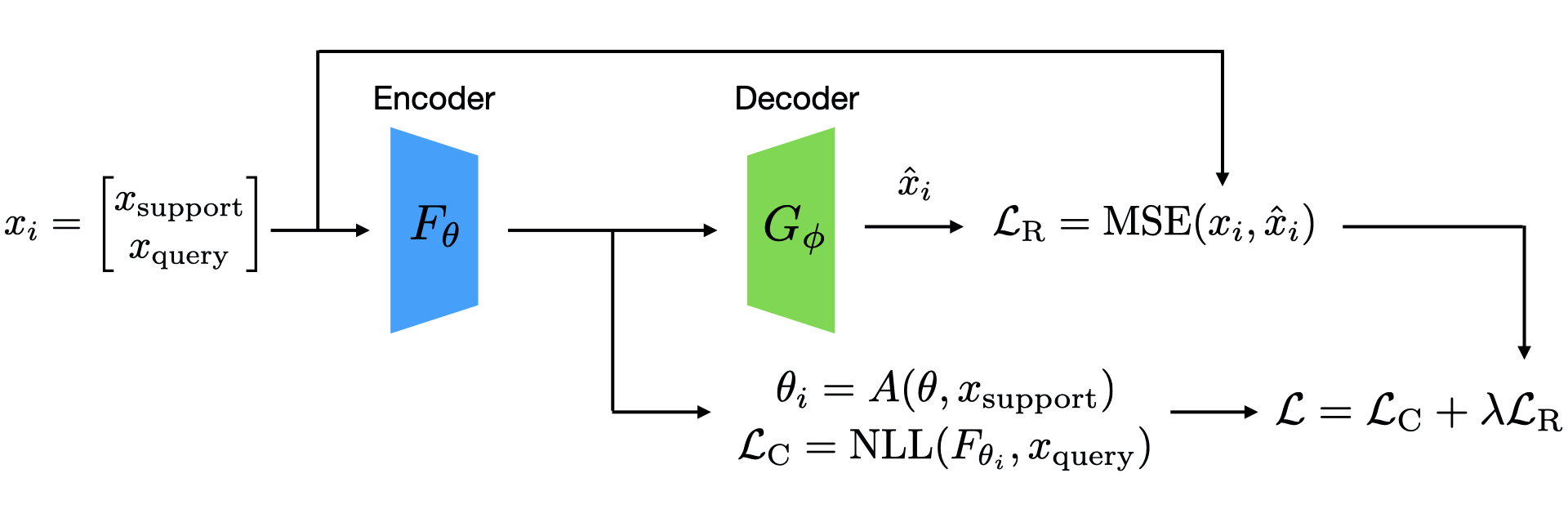}
  \caption{Visualization of the forward pass through AutoProtoNet.}
  \label{fig:autoprotonet-forward}
\end{figure}

AutoProtoNet consists of an encoder-decoder architecture which compresses the input to produce an embedding which must be reconstructed by the decoder. There 4 sequential convolution blocks for the encoder and 4 sequential transpose convolution blocks for the decoder. The details of these blocks can be found in Table~\ref{table:autoprotonet-arch} of Appendix~\ref{appendix:section-architecture}. A forward pass through the model is shown in Figure~\ref{fig:autoprotonet-forward}.

Output padding is used in the second transpose convolution block of the decoder to ensure that the output size of the final transpose convolution block matches the input $84 \times 84$ dimensions of \emph{mini}ImageNet images, but no output padding modifications are necessary for CIFAR-FS images. 

Our architectural design choices imply that a $84 \times 84$ \emph{mini}ImageNet image is embedded as $1600$-dimensional vector, while a $32 \times 32$ CIFAR-FS image is embedded as $256$-dimensional vector.

\subsection{Training}
\label{subsection-model-details}

\begin{algorithm}
\caption{AutoProtoNet Meta-Learning}
\label{alg:autoprotonet-meta-learning}
    \algorithmicrequire Encoder and decoder networks, $F_{\theta}$ and $G_{\phi}$, where $\psi = [\theta; \phi]$ \\
    \algorithmicrequire Fine-tuning algorithm, $A$ \\
    \algorithmicrequire Reconstruction loss weight, $\lambda$ \\
    \algorithmicrequire Learning rate, $\gamma$ \\
    \algorithmicrequire Distribution over tasks, $p(\mathcal{T})$
    \begin{algorithmic}[1]
        \State Initialize $\theta, \phi$, the weights of encoder and decoder 
        \While{not done}
            \State Sample batch of tasks $\{\mathcal{T}_i\}_{i=1}^{n}$, where $\mathcal{T}_i \sim p(\mathcal{T})$ and $\mathcal{T}_i = (\mathcal{T}_i^s, \mathcal{T}_i^q)$
            \For{i=1,...,n}
                \State $\hat{\mathcal{T}_i} \gets  G_{\phi}(F_{\theta}(\mathcal{T}_i))$ \label{line:reconstruct-data}\Comment{Reconstruct task data}
                \State $\mathcal{L}_R \gets \mathrm{MSE}(\mathcal{T}_i, \hat{\mathcal{T}_i })$ \Comment{Compute reconstruction loss}
                \State $\theta_i \gets A(\theta, \mathcal{T}_i^s)$ \label{line:finetune-autoprotonet}\Comment{Compute prototypes (inner loop)} 
                \State $\mathcal{L}_C \gets \mathrm{NLL}(F_{\theta_i}, \mathcal{T}_i^q)$ \Comment{Compute classification loss}
                \State $\mathcal{L} \gets \mathcal{L}_C + \lambda \mathcal{L}_R$
                \State $g_i \gets \nabla_{\psi}\mathcal{L}$ 
            \EndFor
            \State $\psi \gets \psi - \frac{\gamma}{n}\sum_{i}g_i$ \Comment{Update base model parameters (outer loop)}
            
        \EndWhile
    \end{algorithmic}
\end{algorithm}

Training AutoProtoNet is not much different from training a ProtoNet. The main difference is that we augment the meta-training loop with a reconstruction loss to regularize the embedding space and make it suitable for image reconstruction. We display the forward pass through AutoProtoNet in Figure~\ref{fig:autoprotonet-forward} and adapt the meta-learning framwork from Section~\ref{subsection:meta-learning} to describe the meta-training of AutoProtoNet in Algorithm~\ref{alg:autoprotonet-meta-learning}. 

Our ``base'' model now consists of parameters $\psi$ which is a concatenation of encoder network parameters $\theta$ and decoder network parameters $\phi$. In Line~\ref{line:reconstruct-data} of Algorithm~\ref{alg:autoprotonet-meta-learning}, we pass both support and query data from the current task $\mathcal{T}_i$ through the encoder and decoder to produce a reconstruction $\hat{\mathcal{T}_i}$. This reconstruction is then compared to the original data using mean squared error (MSE) loss. The finetuning algorithm in Line~\ref{line:finetune-autoprotonet} of Algorithm~\ref{alg:autoprotonet-meta-learning} is identical to the description in Section~\ref{subsection:meta-learning}, where $\theta_i = ( \{p_k\}_{i=0}^{k}, \theta ) $ is a tuple consisting of a set of prototypes for every class and the encoder network's model parameters. Both of these are used to compute the likelihood of the true labels of our query data as in Equation~\ref{equation:protonet-likelihood}, which is maximized by minimizing the negative log-likelihood (NLL). Finally, the classification loss $\mathcal{L}_C$ and the reconstruction loss $\mathcal{L}_R$ are summed so they can be jointly optimized.

We meta-train ProtoNet and AutoProtoNet on both \emph{mini}ImageNet and CIFAR-FS. To create a prototype reconstruction baseline, we also train two models which make use of ILSVRC 2012 \cite{deng2009imagenet}, which we refer to as ImageNet Autoencoder and ImageNet AutoProtoNet. Note that because \emph{mini}ImageNet is a subset of ILSVRC 2012, the ImageNet models also provide insight into whether more data during pretraining offers any benefit for meta-learning or prototype reconstructions. All training was performed on a single NVIDIA Quadro P6000 from our internal cluster. Training details for each model used in this work are described below.

\paragraph{ProtoNet} Using Algorithm~\ref{alg:meta-learning-framework}, we meta-train a standard ProtoNet for 30 epochs using SGD. Our SGD optimizer uses Nesterov momentum of $0.9$, weight decay of $5 \times 10^{-4}$, and a learning rate of $0.1$, which we decrease to $0.06$ after $20$ epochs.

\paragraph{AutoProtoNet} Using Algorithm~\ref{alg:autoprotonet-meta-learning}, we meta-train an AutoProtoNet for 30 epochs using SGD. We use the same SGD settings as in ProtoNet training. We use a reconstruction loss weight $\lambda = 1$. Following \cite{snell2017prototypical}, both ProtoNet and AutoProtoNet models were trained using 20-way 5-shot episodes, where each class contains $15$ query points per episode, for $30$ epochs. 

\paragraph{ImageNet Autoencoder} We train an autoencoder of the same architecture as AutoProtoNet using only mean squared error (MSE) loss on ILSVRC 2012 \cite{deng2009imagenet} for 20 epochs. We use the SGD optimizer with Nesterov momentum of $0.9$, weight decay of $5 \times 10^{-4}$, and a learning rate of $0.1$, which we decrease by a factor of $10$ every $5$ epochs. To evaluate this model's performance on benchmark few-shot image classification datasets, we make use of the only the encoder to produce embeddings and produce classification labels using the standard ProtoNet classification rule.

\paragraph{ImageNet AutoProtoNet} We use the encoder and decoder weights from the ImageNet Autoencoder as a starting point for the weights of an AutoProtoNet. All other training details are identical to that of AutoProtoNet, which we meta-train using Algorithm~\ref{alg:autoprotonet-meta-learning}.

The 5-way 5-shot test set accuracies of all models used in this work are shown in Table~\ref{table-accuracies}. AutoProtoNet is able to maintain the same level of few-shot image classification accuracy on benchmark datasets as a standard ProtoNet. While we expected AutoProtoNet to have an advantage due to having to incorporate features useful for reconstruction into embeddings, our results suggest that these reconstruction features are not always useful. Given the additional ILSVRC 2012 \cite{deng2009imagenet} data during pretraining, we also expected that ImageNet AutoProtoNet would outperform all other models, but our test results demonstrate that representations learned for image reconstruction are not too helpful for few-shot image classification. Test set accuracies for ImageNet Autoencoder underscore the point that an embedding space trained for only reconstruction is by no means competitive for few-shot classification, though it does achieve better than chance accuracy. 

\section{Experiments}

\subsection{Prototype Visualization}
\label{subsection:prototype-visualizations}

While a standard ProtoNet employs an intuitive nearest-neighbor classification rule for query points, there is no intuitive way for a user to understand what a prototype embedding represents. Prototypical embeddings are crucial to understanding the decision boundaries of ProtoNets. The idea is that a ProtoNet embeds similar images nearby in embedding space, but without a way to visualize these embeddings, we argue that a human practitioner would be unable to debug or improve their deployed model. AutoProtoNet addresses this issue by learning an embedding space that is suitable for image reconstruction.

\begin{figure}%
    \centering
    \subfloat[\centering \emph{mini}ImageNet]{{\includegraphics[width=0.46\textwidth]{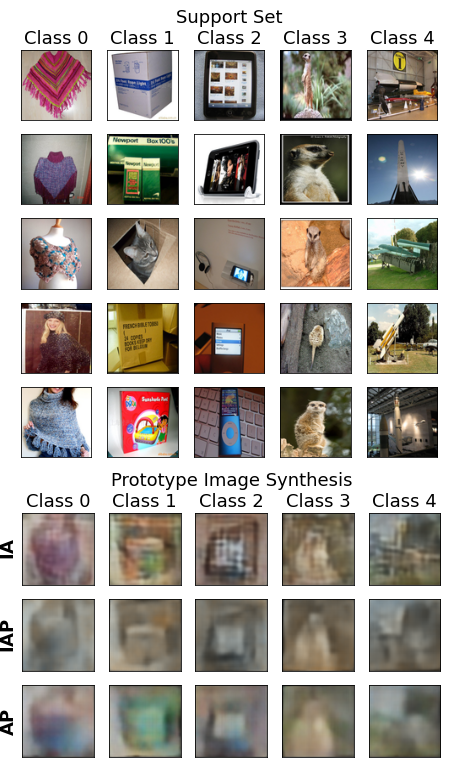} }}
    \qquad
    \subfloat[\centering CIFAR-FS]{{\includegraphics[width=0.46\textwidth]{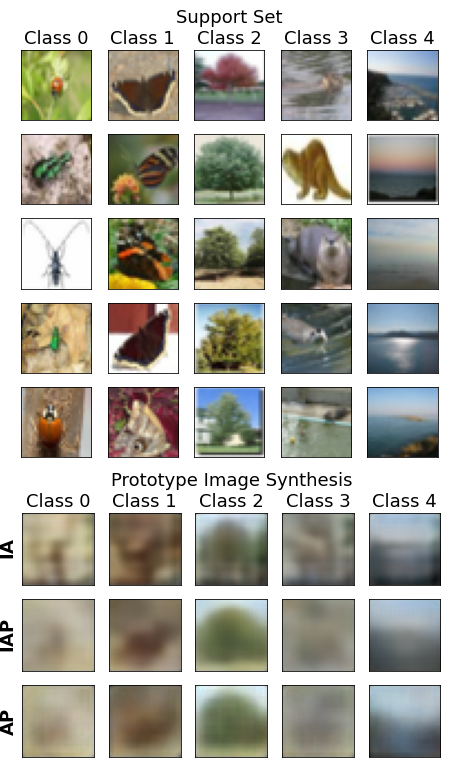} }}
    \caption{Support sets for a 5-way 5-shot validation task of \emph{mini}ImageNet (a) and CIFAR-FS (b). The embeddings of every image within a class are averaged to form a prototype embedding which is then synthesized as an image by using the decoder of an ImageNet Autoencoder (\textbf{IA}), an ImageNet AutoProtoNet (\textbf{IAP}), and an AutoProtoNet (\textbf{AP}). }
    \label{fig:prototypes}
\end{figure}

Figure~\ref{fig:prototypes} displays prototype visualizations given a validation support set from \emph{mini}ImageNet and CIFAR-FS. The ImageNet Autoencoder (\textbf{IA}) and ImageNet AutoProtoNet (\textbf{IAP}) were both pretrained on all of ILSVRC 2012 \cite{deng2009imagenet}, and so classes present in this validation support set are not novel classes because \emph{mini}ImageNet is a subset of ILSVRC 2012. However, in the case of the AutoProtoNet (\textbf{AP}), the classes in this validation support set are novel and the synthesized prototype images remain qualitatively on-par with the models trained with more data (such as ImageNet Autoencoder), suggesting that meta-tasks during training were sufficient to regularize an embedding space suitable for image synthesis. Analyzing the prototype reconstructions from CIFAR-FS in Figure~\ref{fig:prototypes}(b), we see that prototype visualizations are generally too blurry to help a human determine whether the model has learned a sufficient representation of a class. We believe part of the problem is the low resolution and size of CIFAR-FS images.

\begin{table}
  \caption{5-way 5-shot test set accuracies with 95\% confidence intervals.} 
  \label{table-accuracies}
  \centering
  \begin{tabular}{lcc}
    \toprule
    Model & \emph{mini}ImageNet & CIFAR-FS \\
    \midrule
    ImageNet Autoencoder & $36.83 \pm 0.48$\% & $46.08 \pm 0.58$\% \\
    ImageNet AutoProtoNet & $70.76 \pm 0.51$\% & $79.65 \pm 0.52$\% \\
    \midrule
    ProtoNet & $70.20 \pm 0.52$\% & $80.31 \pm 0.51$\% \\
    AutoProtoNet & $70.61 \pm 0.52$\% & $80.16 \pm 0.52$\% \\
    \bottomrule \\
  \end{tabular}
\end{table}

\subsection{Human-guided Prototype Refinement}
\label{subsection:human-guided-prototype-refinement}

\begin{figure}%
    \centering
    \subfloat[\centering Support set and prototype visualizations]{{\includegraphics[width=0.45\textwidth]{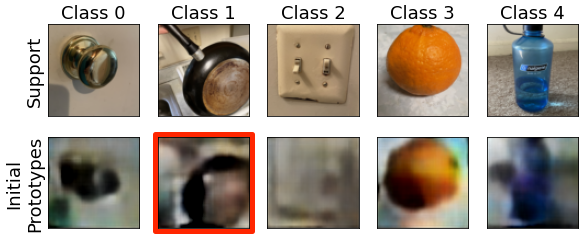} }}
    \qquad
    \subfloat[\centering New image and corresponding embedding]{{\includegraphics[width=0.45\textwidth]{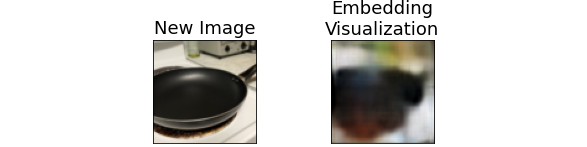} }}
    \qquad
    \subfloat[\centering Interpolating 10 steps from initial prototype to new image embedding]{{\includegraphics[width=\textwidth]{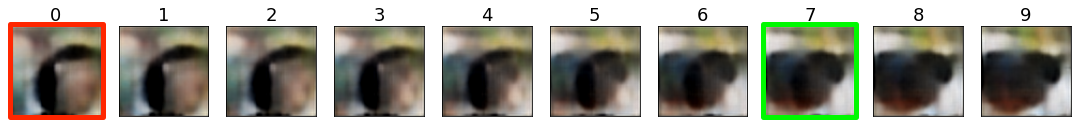} }}
    \qquad
    \subfloat[\centering New set of prototypes]{{\includegraphics[width=0.5\textwidth]{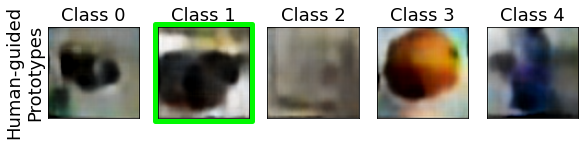} }}
    \caption{Steps for human-guided prototype selection in a 5-way 1-shot task. Step (a): a human chooses an \textcolor{red}{initial prototype} to refine. Step (b): a human captures one additional image to guide prototype refinement. Step (c): Interpolations between the \textcolor{red}{initial prototype} and the new image embedding (index 9) are shown to the human and a \textcolor{green}{new prototype} selection is made. Step (d): A new set of prototypes is set, with class 2 having been refined.}
    \label{fig:human-guided-prototypes}
\end{figure}

%
%
%

To highlight the benefits of an embedding space suitable for image reconstruction, we designed an experiement to demonstrate how a human can guide prototype selection at test-time using AutoProtoNet. Assuming the user knows the kinds of images the model will encounter at inference time and given the ability to capture one more image, could we refine an initial prototype to achieve higher accuracy on the validation set?

\paragraph{Data Collection}

Based on objects we had around the house, we chose to formulate a 5-way 1-shot classification problem between ``door knob'', ``frying pan'', ``light switch'', ``orange'', and ``water bottle''. Note that ``orange'' and ``frying pan'' are classes in the \emph{mini}ImageNet training split, but all other classes are novel. Because we sought to demonstrate how one might use an AutoProtoNet in a real-world setting, all $55$ images in this task are novel, in-the-wild images, captured using an iPhone 12. Our support set consists of $5$ images ($1$ image per class). Our validation set consists of $50$ images ($10$ images per class) and can be found in Figure~\ref{fig:prototype-refinement-test-set} of Appendix~\ref{appendix:section-validation-set}. 

\paragraph{Prototype Refinement} 


Prototype refinement is a debugging technique meant for cases in which a human believes prototype visualization may not be representative of the class. To exaggerate the idea of prototype refinement, we purposefully choose the back-side of a frying pan as a support image for class 1 (``frying pan'') so that the prototype visualization has undesirable image features. Generally, a prototype for an arbitrary object of a novel class is likely to be visually ambiguous if the embedding network did not train on a suitable dataset, so this setup is conceivable in the real-world. 

For our classification model, we make use of the AutoProtoNet described in Section~\ref{subsection-model-details}. To apply AutoProtoNet to this new classification task, we ``fine-tine'' AutoProtoNet by providing a support set shown in Figure~\ref{fig:human-guided-prototypes}(a). After meta-learning, an AutoProtoNet's only changeable parameters are its prototypes which, by design, can be reconstructed into images using the decoder. By visually understanding an AutoProtoNet's embedding space, a user can choose to change image features of a prototype reconstruction, thus changing the prototype itself. In contrast, a standard ProtoNet performs inference using its support data, which is visually inaccessible and uninterpretable. 

Using a newly captured image $x \in \mathbb{R}^{d}$, we use the encoder $F_{\theta}$ to generate an embedding $p = F_{\theta}(x)$. Given an initial prototype $p_k$ for class $k$, we use the decoder $G_{\phi}$ to synthesize images $\hat{x}_i \in \mathbb{R}^{d}$ for interpolations between $p_k$ and $p$ as follows:

\begin{equation}
    \hat{x}_i = G_{\phi}( (1-\alpha) p_k + \alpha p ) \qquad \alpha \in [0,1]
\end{equation}

\paragraph{Results}

Using the initial prototypes from Figure~\ref{fig:human-guided-prototypes}(a), AutoProtoNet achieves $80\%$ accuracy on the validation set consisting of $50$ images from all $5$ classes. The $10$ misclassified images are all of the ``frying pan'' class. After debugging the ``frying pan'' prototype by capturing an additional image of a correctly-oriented frying pan and choosing an interpolation, the resulting embedding is used as the new support as shown in Figure~\ref{fig:human-guided-prototypes}(d). Under the new human-guided prototypes, AutoProtoNet achieves an accuracy of $98\%$ on the validation set, where the single misclassified image is of the ``door knob'' class.

The novelty of our method lies in the ability for a human to fine-tune the model in an interactive way, leading to a performance increase in validation set accuracy. In this example, AutoProtoNet's decoder allowed for the visualization of the prototype embedding, which we found to be visually incorrect. Thus, we captured an additional, more representative image to designate the direction in which to move the initial prototype to fit a human-designated criteria. 



\section{Conclusion}
\label{section:conclusion}
With AutoProtoNet, we present a step toward meta-learning approaches capable of giving some insight into their learned parameters. We argue that if meta-learning approaches are to be useful in practice, there should be ways for a human to glean some insight into why a classification might have been made. Through prototype visualizations and a prototype refinement method, we highlight the benefits of AutoProtoNet and take steps to improve a simple few-shot classification algorithm by making it more interpretable while maintaining the same degree of accuracy as a standard ProtoNet.


Our proposed method could likely be extended to Relation Networks \cite{sung2017relationnets}, MetaOptNet \cite{lee2019metaoptnet}, or R2D2 \cite{bertinetto2018meta}, with a decoder network to visualize embeddings. It may also be possible to meta-train a variational autoencoder to learn a latent space more suitable for detailed image synthesis. We believe generative models can play a larger role in interpretability of meta-learning algorithms.

To confirm the effectiveness of our interpretability results, we intend to perform a human subjects study where a human determines whether prototype visualizations help in understanding classification results. We also recognize the limits of using a small dataset to evaluate the performance of our prototype refinement method. We leave the creation of a larger, more diverse validation set to future work.

\begin{ack}
This work was supported by the Office of Naval Research (WL). The views and conclusions contained in this document are
those of the authors and should not be interpreted as necessarily representing the official policies, either expressed or implied, of the U.S. Navy.
\end{ack}

\bibliography{references}
\bibliographystyle{abbrvnat}

\newpage

\appendix

\section{Validation Set for Custom Classification Task}
\begin{figure}[!h]
    \centering
    \includegraphics[width=\textwidth]{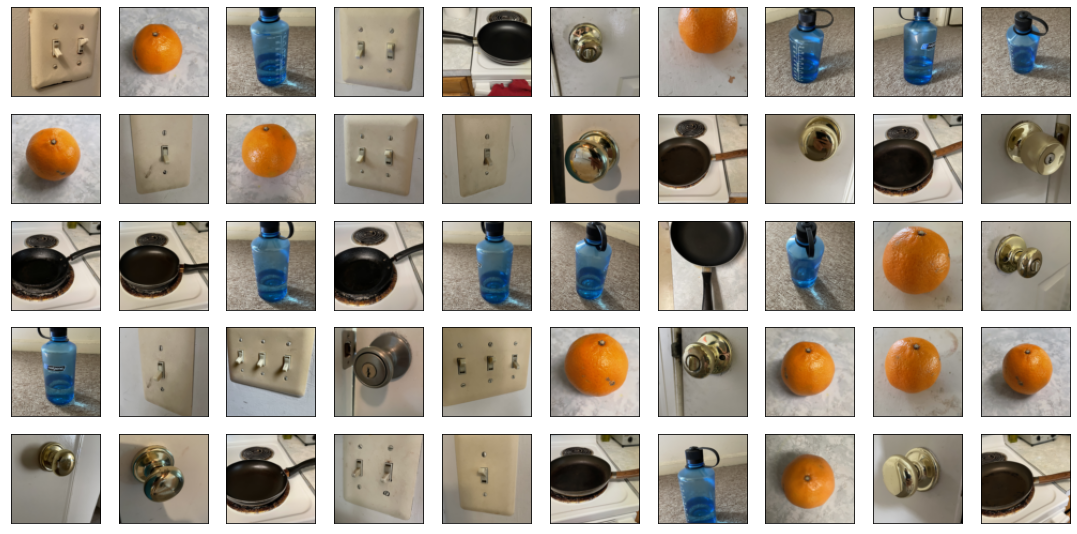}
    \caption{Validation set for experiment described in Section~\ref{subsection:human-guided-prototype-refinement}}
    \label{fig:prototype-refinement-test-set}
\end{figure}

In Figure~\ref{fig:prototype-refinement-test-set}, we display the $50$ images of our custom $5$-way validation set. The images from the ``light switch'' and ``door knob'' classes are diverse in terms of shape, pose, and lighting condition.

\label{appendix:section-validation-set}

\section{Architecture Details}
\label{appendix:section-architecture}

In our description of the AutoProtoNet architecture in Table~\ref{table:autoprotonet-arch}, we display output sizes for the first Conv Block of the encoder and the first Conv Transpose Block of the decoder, assuming an $84 \times 84$ \emph{mini}ImageNet image is used as input. 

\begin{table}[!h]
  \centering
  \caption{AutoProtoNet Architecture Components}
  \begin{tabular}{ccccccc}
    \toprule
    \multicolumn{3}{c}{Conv Block} & \multicolumn{1}{c}{ } & \multicolumn{3}{c}{Conv Transpose Block} \\
    \cmidrule(l){1-3} \cmidrule(r){5-7} \\
    Layer      & Parameters         & Output Size              & & Layer     & Parameters     & Output Size\\
    \midrule
    Conv       & $3 \times 3$, $64$ & $64 \times 84 \times 84$ & & Conv Transpose & $2 \times 2$, $*2$ & $64 \times 10 \times 10$ \\
    Batch Norm &                    & & & Batch Norm     & &  \\
    Max Pool   & $3 \times 3$, /2  &  $64 \times 42 \times 42$ & & Conv           & $3 \times 3$, $64$ & $64 \times 10 \times 10$ \\
    \bottomrule \\
  \end{tabular}
  \label{table:autoprotonet-arch}
\end{table}

\section{Implementation Details}
\label{appendix:implementation}

We use PyTorch \cite{paszke2019pytorch} and work on a fork of code used for \cite{goldblum2019robust}, which uses the MIT License. Our fork can be used to reproduce experiments and is available here: \url{https://github.com/psandovalsegura/AdversarialQuerying}.

\end{document}